\documentclass[sigplan]{acmart}
\renewcommand\footnotetextcopyrightpermission[1]{}
\usepackage{mathrsfs}
\usepackage{graphicx}
\usepackage{minted}
\usepackage{xspace}
\usepackage{xcolor}
\usepackage{enumitem}
\usepackage{multirow}
\usepackage{booktabs}

\setlist{nosep}

\acmConference{}{}{}{}
\acmBooktitle{}

\title{AutoRAGTuner: A Declarative Framework for Automatic Optimization of RAG Pipelines}

\author{Xintan Zeng, Yongchao Liu, Yice Luo, Jiajun Zheng}
\affiliation{%
  \institution{Ant Group, China}
  \country{}
}
\email{{xintan.zxt, yongchao.ly, luoyice.lyc}@antgroup.com, zhengjiajun@gmail.com}

\begin{document}

\maketitle
Retrieval-Augmented Generation (RAG) enhances LLMs, but performance is highly sensitive to complex architecture designs and hyper-parameter configurations, which currently rely on inefficient manual tuning. We present \textbf{AutoRAGTuner}, a declarative, configuration-driven framework that automates the RAG life-cycle: construction, execution, evaluation, and optimization. AutoRAGTuner employs a modular architecture to decouple pipeline stages through a component registration mechanism. To unify heterogeneous data, we introduce the \textbf{Domain-Element Model (DEM)}, representing objects as atomic elements with bidirectional pointers to support nodes, edges, and hyperedges. Furthermore, AutoRAGTuner integrates an adaptive \textbf{Bayesian optimization engine} for end-to-end hyper-parameter tuning. Experimental results demonstrate AutoRAGTuner’s architectural generality: across diverse RAG pipelines—ranging from vanilla to graph-based—the framework consistently outperforms default baselines. Notably, AutoRAGTuner significantly mitigates engineering overhead, where its declarative configuration language enables a up to 95\% reduction in code churn for architectural adjustments. Overall, AutoRAGTuner provides a systematically optimizable foundation for building evolvable and reusable RAG systems.
\section{Background and Motivation}
Retrieval-Augmented Generation (RAG) improves the quality of LLM answer by leveraging external knowledge bases~\cite{rag} . However, the performance of a typical RAG pipeline is highly sensitive to its structure and hyper-parameters. Manual tuning faces two primary challenges: (1) a single modification often forces upstream and downstream code to recouple, requiring re-debugging of the entire pipeline, resulting in low efficiency; (2) there is no unified, declarative way to define pipeline structure and hyper-parameters, making them difficult to standardize, reuse, or reproduce.

Recent work on automated RAG optimization, such as AutoRAG~\cite{autorag} , employs greedy algorithms to optimize hyper-parameters across various modules, including query expansion, retrieval, and generation. AutoRAG-HP~\cite{autorag-hp} models hyper-parameter selection as a hierarchical multi-armed bandit problem to achieve more flexible online hyper-parameter tuning. However, these frameworks treat RAG pipelines as fixed structures. This structural rigidity hinders native support for more sophisticated retrieval strategies, such as the complex graph topologies required for multi-hop reasoning. Moreover, any architectural adjustment typically requires invasive code refactoring.

In contrast, AutoRAGTuner enables hyper-parameter exploration within a flexible architectural search space via declarative orchestration. By unifying heterogeneous data modeling, it supports diverse retrieval strategies, filling the gap in architectural flexibility and strategy diversity while providing a standardized, replayable, and optimizable framework for RAG optimization.
\begin{figure}[t!]
  \centering
  \includegraphics[width=\columnwidth]{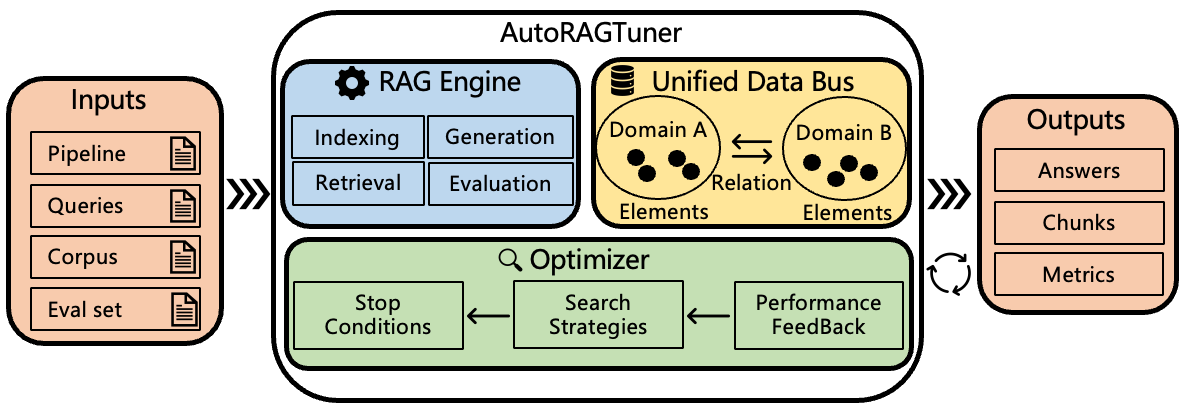}
  \caption{System overview of AutoRAGTuner.}
  \label{fig:autoragtuner-arch}
\end{figure}
\section{System Implementation}

As shown in Figure~\ref{fig:autoragtuner-arch}, AutoRAGTuner is built on a modular architecture where pipeline stages are decoupled via a component registration mechanism. All components follow a unified interface contract and can be implemented either in C++ or Python, enabling a practical trade-off between performance and rapid feature development.

\begin{figure}[t]
  \centering
  \includegraphics[width=\columnwidth]{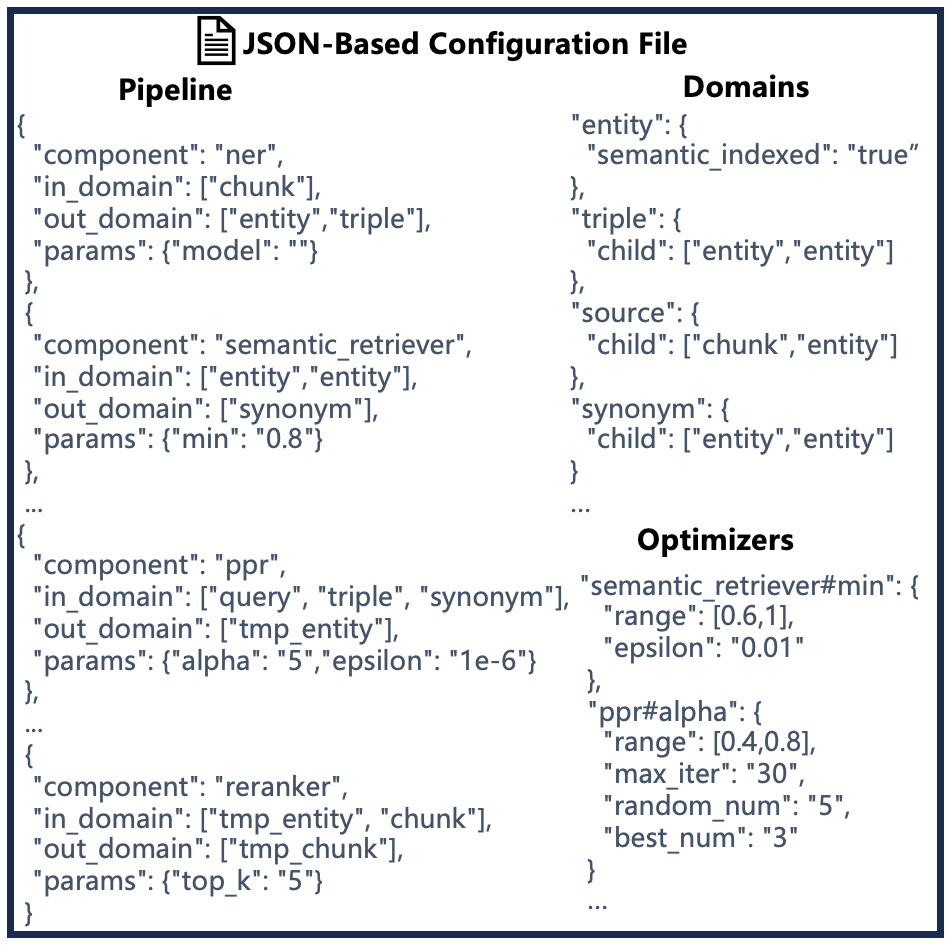}
  \caption{An example JSON configuration for declarative orchestration in AutoRAGTuner.}
  \label{fig:json-config}
\end{figure}

To enable an "Edit-and-Run" paradigm without recompilation, AutoRAGTuner provides a \textbf{declarative JSON orchestration language} (Figure~\ref{fig:json-config}). Developers specify the pipeline composition and optimization strategies through configuration fragments, allowing structural adjustments without modifying the underlying codebase.

To minimize stage coupling, AutoRAGTuner introduces the \textbf{Domain-Element Model} (DEM). DEM abstracts heterogeneous retrieval objects (e.g., text chunks, entities, relations) as \textbf{Atomic Elements}. Each element contains basic attributes (e.g., ID, weight) and an extensible property set. DEM introduces bidirectional pointers, allowing an element to function as an independent node or a container for multiple children to form edges or multidimensional hyperedges. These relations are declared via the Domain configuration in a JSON specification (e.g., declaring an entity element as a child of a chunk element), and materialized by corresponding components as edges stored in DEM. This unified representation supports both hierarchical structures and complex graph topologies within a single abstraction.

DEM uses \textbf{Domain} as a logical partition to group and manage elements. Through JSON declarations, developers can define relationships across domains and trigger domain-specific vector indexing for semantic retrieval. At runtime, components share data via a unified data bus to maintain end-to-end consistency and life-cycle management. Thus, component-level changes do not require upstream or downstream refactoring.

For optimization, AutoRAGTuner integrates an adaptive \textbf{Bayesian autotuning engine} for end-to-end optimization. It employs a hybrid strategy, starting with random exploration to establish a prior, followed by exploitation guided by an acquisition function based on \textbf{Expected Improvement} to focus on high-potential regions. To minimize computational costs, the engine supports epsilon-convergence and maximum iteration limits, alongside warm-start capabilities to reuse previous observation traces.

\section{Preliminary Results}We assess AutoRAGTuner based on three criteria: Recall@5 (R@5), $F_1$ score, and tuning efficiency. To demonstrate architectural generality, we implement two representative RAG pipelines: a standard \textbf{Vanilla RAG} based on the "retrieve-and-generate" paradigm, and a complex \textbf{Graph RAG}~\cite{peng2025graph} named HippoRAG~\cite{hipporag}. Our experiments employ Kimi-K2-Instruct-0905 as the LLM and Qwen3-Embedding-4B as the retriever.

For Vanilla RAG, we declare two key hyper-parameters for automated optimization: the chunk size and the overlap ratio during the indexing phase, with the unchunked configuration as the baseline. For HippoRAG, we declare three key hyper-parameters accordingly: the cosine similarity threshold for synonym relationship construction during indexing, the cosine similarity threshold for linking similar entities during retrieval, and the damping factor in the Personalized PageRank (PPR) algorithm. The baseline of HippoRAG follows the official settings from its original paper. Optimization is conducted using 200 HotPotQA~\cite{hotpotqa} training samples, with evaluation on 1,000 development samples each from HotPotQA and 2WikiMultiHopQA~\cite{2wiki}.

\begin{figure}[t]
  \centering
  \includegraphics[width=\columnwidth]{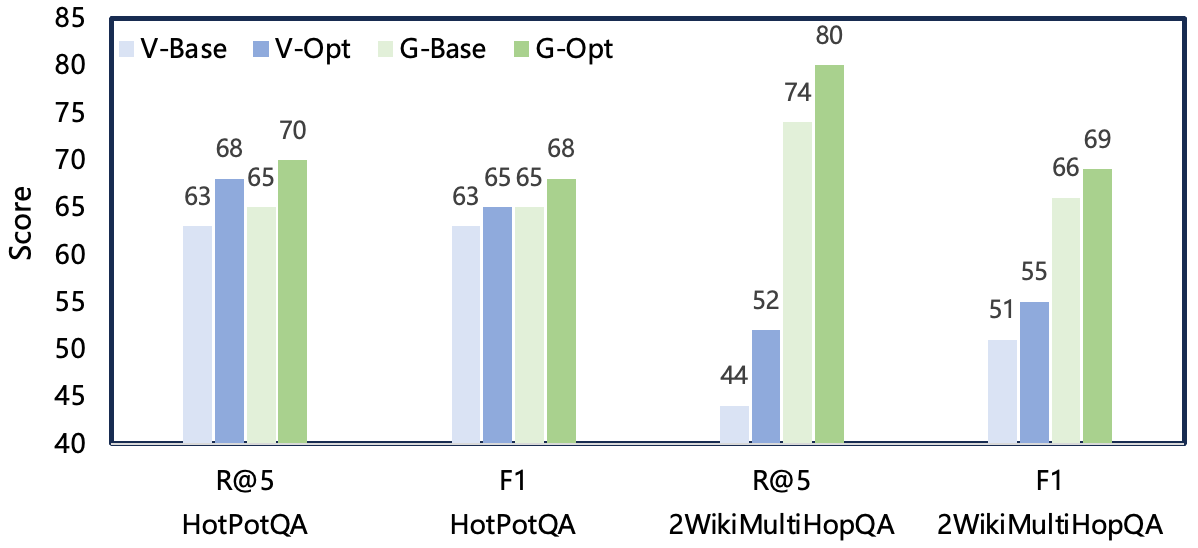}
  \caption{Performance improvement with AutoRAGTuner. 
V-Base/V-Opt refer to the baseline and optimized Vanilla RAG, with G-Base/G-Opt to the baseline and optimized HippoRAG.}
  \label{fig:performance}
\end{figure}

Figure~\ref{fig:performance} underscores AutoRAGTuner’s architectural generality and the efficacy of its automated optimization. Across diverse pipelines, the framework achieves robust gains of 5\% to 8\% in R@5, with $F_1$ scores improving by up to 4\%. Beyond quality improvements, Table~\ref{tab:efficiency} highlights the agility of AutoRAGTuner to support architectural evolution. Unlike manual tuning, which requires invasive code refactoring, AutoRAGTuner abstracts pipeline orchestration into a declarative, configuration-driven layer. Using unified data abstractions to isolate structural dependencies, the framework effectively minimizes engineering overhead. In conclusion, AutoRAGTuner provides a robust foundation for evolvable and systematically optimizable RAG pipelines.

\begin{table}[t]
  \centering
  \footnotesize
  \caption{Efficiency Comparison.}
  \label{tab:efficiency}
  \begin{tabular}{lcc}
    \toprule
    Method & Code Changes (LOC) & Time (Debug + Tuning) \\
    \midrule
    Manual Tuning & $\ge$1000 & 1 Day + 2 Weeks \\
    AutoRAGTuner & $\le$50 & $\approx$ 10 Hours \\
    \bottomrule
  \end{tabular}
\end{table}

\bibliographystyle{unsrt}
\bibliography{refs}

\end{document}